\documentclass{article}

\PassOptionsToPackage{numbers}{natbib}
\usepackage[accepted]{icml2020}

\usepackage[utf8]{inputenc} 
\usepackage[T1]{fontenc}    
\usepackage{booktabs}       
\usepackage{amsfonts}       
\usepackage{nicefrac}       
\usepackage{microtype}      
\usepackage{xcolor}
\usepackage{graphicx}
\usepackage{amsmath}
\usepackage{placeins}
\usepackage{subfigure}
\usepackage{mathtools}
\usepackage[utf8]{inputenc}
\usepackage{subfigure}
\usepackage{multirow}
\usepackage{color,verbatim,caption}
\PassOptionsToPackage{hyphens}{url}\usepackage[hidelinks]{hyperref}
\usepackage{enumitem}
\usepackage{dsfont}
\usepackage{balance}
\usepackage{textcomp}
\usepackage{stfloats}

\usepackage{array}
\newcolumntype{L}[1]{>{\raggedright\let\newline\\\arraybackslash\hspace{0pt}}m{#1}}
\newcolumntype{C}[1]{>{\centering\let\newline\\\arraybackslash\hspace{0pt}}m{#1}}
\newcolumntype{R}[1]{>{\raggedleft\let\newline\\\arraybackslash\hspace{0pt}}m{#1}}

\newcommand{\mat}[1]{\boldsymbol{#1}}
\newcommand{\der}[0]{\mathrm{d}}
\newtheorem{proposition}{Proposition}
\newtheorem{lemma}{Lemma}

\icmltitlerunning{Continuous Graph Neural Networks}

\begin{document}
\twocolumn[
\icmltitle{Continuous Graph Neural Networks}



\icmlsetsymbol{equal}{*}

\begin{icmlauthorlist}
\icmlauthor{Louis-Pascal A.~C.~Xhonneux}{equal,mila,udem}
\icmlauthor{Meng Qu}{equal,mila,udem}
\icmlauthor{Jian Tang}{mila,hec,cifar}
\end{icmlauthorlist}

\icmlaffiliation{mila}{Mila - Quebec AI Institute, Montr\'eal, Canada}
\icmlaffiliation{udem}{University of Montr\'eal, Montr\'eal, Canada}
\icmlaffiliation{hec}{HEC Montr\'eal, Montr\'eal, Canada}
\icmlaffiliation{cifar}{CIFAR AI Research Chair}

\icmlcorrespondingauthor{Jian Tang}{jian.tang@hec.ca}

\icmlkeywords{Machine Learning, ICML}

\vskip 0.3in
]



\printAffiliationsAndNotice{\icmlEqualContribution with order determined by flipping a coin.} 

\begin{abstract}

This paper builds on the connection between graph neural networks and traditional dynamical systems.
We propose \emph{continuous graph neural networks} (CGNN), which generalise existing graph neural networks with discrete dynamics in that they can be viewed as a specific discretisation scheme. The key idea is how to characterise the continuous dynamics of node representations, i.e.\ the derivatives of node representations, w.r.t.\ time. Inspired by existing diffusion-based methods on graphs (e.g.\ PageRank and epidemic models on social networks), we define the derivatives as a combination of the current node representations, the representations of neighbors, and the initial values of the nodes. We propose and analyse two possible dynamics on graphs---including each dimension of node representations (a.k.a.\ the feature channel) change independently or interact with each other---both with theoretical justification. The proposed continuous graph neural networks are robust to over-smoothing and hence allow us to build deeper networks, which in turn are able to capture the long-range dependencies between nodes. Experimental results on the task of node classification demonstrate the effectiveness of our proposed approach over competitive baselines.

\end{abstract}

\section{Introduction}

Graph neural networks (GNNs) have been attracting growing interest due to their simplicity and effectiveness in a variety of applications such as node classification~\cite{kipf16, veli17}, link prediction~\cite{zhang2018link}, chemical properties prediction~\cite{gilmer2017neural}, and natural language understanding~\cite{marcheggiani2017encoding, yao2019graph}. The essential idea of GNNs is to design multiple graph propagation layers to iteratively update each node representation by aggregating the node representations from their neighbours and the representation of the node itself. In practice, a few layers (two or three) are usually sufficient for most tasks~\cite{qu2019}, and more layers may lead to inferior performance~\cite{kipf16, zhou2018graph, liqi18}. 

A key avenue to improving GNNs is being able to build deeper networks to learn more complex relationships between the data and the output labels. The GCN propagation layer smooths the node representations, i.e.\ nodes near each other in the graph become more similar~\cite{kipf16}. This can lead to over-smoothing as we stack more and more layers, meaning that nodes representations converge to the same value leading to worse performance \cite{kipf16, zhou2018graph, liqi18}. Thus, it is important to alleviate the node over-smoothing effect, whereby node representations converge to the same value.

Furthermore, it is crucial to improve our theoretical understanding of GNNs to enable us to characterise what signals from the graph structure we can learn. Recent work on understanding GCN \cite{oono2020} has considered GCN as a discrete dynamical system defined by the discrete layers. In addition, \citet{chen18} demonstrated that the usage of discrete layers is not the only perspective to build neural networks. They pointed out that discrete layers with residual connections can be viewed as a discretisation of a continuous Ordinary Differential Equation (ODE). They showed that this approach is more memory-efficient and is able to model the dynamics of hidden layers more smoothly. 

We use the continuous perspective inspired by diffusion based methods to propose a new propagation scheme, which we analyse using tools from ordinary differential equations (i.e.\ continuous dynamical systems). Indeed we are able to explain what representations our model learns as well as why it does not suffer from the common over-smoothing problem observed in GNNs. Allowing us to build `deeper' networks in the sense that our model works well with large values of time. The key factor for the resilience to over-smoothing is the use of a restart distribution as originally proposed in Pagerank~\cite{page1999pagerank} in a continuous setting. The intuition is that the restart distribution helps to not forget the information from low powers of the adjacency matrix, hence enabling the model to converge towards a meaningful stationary distribution.

The main contributions of this paper are:
\begin{enumerate}
    \item We propose two continuous ODEs of increasing model capacity inspired by PageRank and diffusion based methods;
    \item We theoretically analyse the representation learned by our layer and show that as $t\to\infty$ our method approaches a stable fixed point, which captures the graph structure together with the original node features. Because we are stable as $t\to\infty$ our network can have an infinite number of `layers' and is able to learn long-range dependencies;
    \item We demonstrate that our model is memory efficient and is robust to the choice of $t$. Further, we demonstrate on the node classification against competitive baselines that our model is able to outperform many existing state-of-the-art methods.
\end{enumerate}
\section{Preliminaries}

Let a graph $G=(V,E)$\footnote{Throughout the paper we will only consider simple graphs} be defined by vertices $V$ and edges $E\subseteq V\times V$ between vertices in $V$. It is common in graph machine learning to use an adjacency matrix $\mat{Adj}$ as an alternative characterisation. Given a node ordering $\pi$ and a graph $G$, the elements of the adjacency matrix $\mat{Adj}^{|V|\times |V|}$ are defined by the edge set $E$:
\begin{equation}
\mat{Adj}_{ij} = \begin{cases}
1&\quad\mathrm{if}\: (v_i,v_j)\in E\\
0&\quad \mathrm{otherwise}
\end{cases}
\end{equation}
As the degree of nodes can be very different, we typically normalize the adjacency matrix as $\mat{D}^{-\frac 12 }\mat{Adj}\mat{D}^{-\frac 12}$, where $\mat{D}$ is the degree matrix of $\mat{Adj}$. Such a normalized adjacency matrix always has an eigenvalue decomposition, and the eigenvalues are in the interval $[-1, 1]$~\cite{chun97}. The negative eigenvalues can make graph learning algorithms unstable in practice, and hence we follow~\citet{kipf16} and leverage the following regularized matrix for characterizing graph structures:
\begin{equation}
\label{eq: def A}
\mat{A} \coloneqq \frac{\alpha}{2} \left(\mat{I} + \mat{D}^{-\frac 12 }\mat{Adj}\mat{D}^{-\frac 12}\right),
\end{equation}
where $\alpha \in (0,1)$ is a hyperparameter, and the eigenvalues of $\mat{A}$ are in the interval $[0, \alpha]$.

Given such a matrix $\mat{A} \in \mathbb{R}^{|V| \times |V|}$ to characterise the graph structure, and a node feature matrix $\mat{X} \in \mathbb{R}^{|V| \times |F|}$, with $|F|$ being the number of node features, our goal is to learn a matrix of node representations $\mat{H} \in \mathbb{R}^{|V| \times d}$, where $d$ is the dimension of representations, and the $k$-th row of $\mat{H}$ is the representation of the $k$-th node in an ordering $\pi$.

A continuous ODE throughout this paper will refer to an equation of the following form:
\begin{equation}
    \frac{\der \mat{x}}{\der t} = f(\mat{x}(t), t),
\end{equation}
where $x$ may be scaler, vector valued, or matrix valued and $f$ is function that we will parametrise to define the hidden dynamic. \citet{chen18} showed how we can backpropagate through such an ODE equation and hence use it as a building block for a neural network.

\section{Related Work}
\textbf{Neural ODE}
Neural ODE \cite{chen18} is an approach for modelling a continuous dynamics on hidden representation, where the dynamic is characterised through an ODE parameterised by a neural network. However, these methods can only deal with unstructured data, where different inputs are independent. Our approach extends this in a novel way to graph structured data.

\textbf{GNNs}. Graph neural networks~\cite{kipf16,veli17} are an effective approach for learning node representations in graphs. Typically, GNNs model discrete dynamics of node representations with multiple propagation layers, where in each layer the representation of each node is updated according to messages from neighbouring nodes. The majority of GNNs learn the relevant information from the graph structure $\mat{A}$ by learning finite polynomial filters $g$ to apply to the eigenvalues $\mat{\Lambda}$ of the graph Laplacian $\mat{L} = \mat{P}\mat{\Lambda}\mat{P}^{-1}$ in each propagation layer \cite{kipf16, deff16, wije19, veli17}.  GCN \cite{kipf16}, for instance, uses a first-order Chebyshev polynomial. However, existing GNNs (e.g.\ GCN) have been shown \cite{liqi18, zhou2018graph, oono2020} to suffer from over-smoothing, i.e.\ node representations start to converge to the same value. Compared to these studies, we follow continuous dynamical systems to model the continuous dynamic on node representations. Moreover, our approach does not have the over-smoothing problem,  because our model converges to a meaningful representation (fixed point) as $t\to\infty$.

A proposed solution in recent work is to either add a residual connection to the preceding layer \cite{avel2019} or to use a concatenation of each layer \cite{wije19, luan19, xu2018, dehm19}. The latter approach does not scale to very deep networks due to the growing size of the representation. \citet{chen18}'s approach of turning residual connections from the preceding layer into a continuous ODE has gathered much attention \cite{deng19, avel2019, zhua2020, poli2019}. In contrast to many of these works we are able to provide a theoretical justification for our ODE and our representation does not grow with depth.

A significant amount of work has focused on understanding the theoretical properties of GCN and related architectures better \cite{oono2020, dehm19, nt2019}. \citet{dehm19} argue that to learn the topology of a graph the network must learn the graph moments---an ensemble average of a polynomial of $\mat{Adj}$. They show that GCN can only learn graph moments of a specific power of $\mat{Adj}$. To remedy this they concatenate $[\mat{h}^{(1)}, \mat{h}^{(2)},\ldots, \mat{h}^{(n)}]$ the feature maps (output) of each layer. \citet{oono2020} demonstrate that deeper GCNs exponentially lose expressive power by showing that in the limit as the number of layers goes to infinity the node representation is projected onto the nullspace of the Laplacian, which implies that two nodes with identical degree in the same connected component are will have the same representation. 
Instead we propose a continuous dynamical system which does not suffer from this problem and demonstrate so as $t\to\infty$. \citet{nt2019} focus on explaining which assumptions hold such that GCN \cite{kipf16} and SGC \cite{wu2019} work on the common citation networks. Our work fits into the existing literature by using the Neural ODE framework to provide a novel intuition to what is needed to address loss of expressive power with depth. We do this by proposing and analysing our specific solution.

\textbf{Concurrent work}.
Several concurrent works have developped similar ideas; \citep{poli2019} proposes an ODE based on treating the GCN-layer as a continuous vector field and combines discrete and continuous layers. Other concurrent works \cite{deng19, zhua2020} use the Neural ODE framework and parametrise the derivative function using a 2- or 3-layer GNN directly, instead we develop a continuous message-passing layer and do not use a discrete deep neural network to parametrise the derivative. In contrast, we motivate our ODE from diffusion based methods and theoretically justify how our approach helps to address over-smoothing as well as long-range connections between nodes. 
\section{Model}

\begin{figure*}[t]
    \centering
    \includegraphics[width=\linewidth]{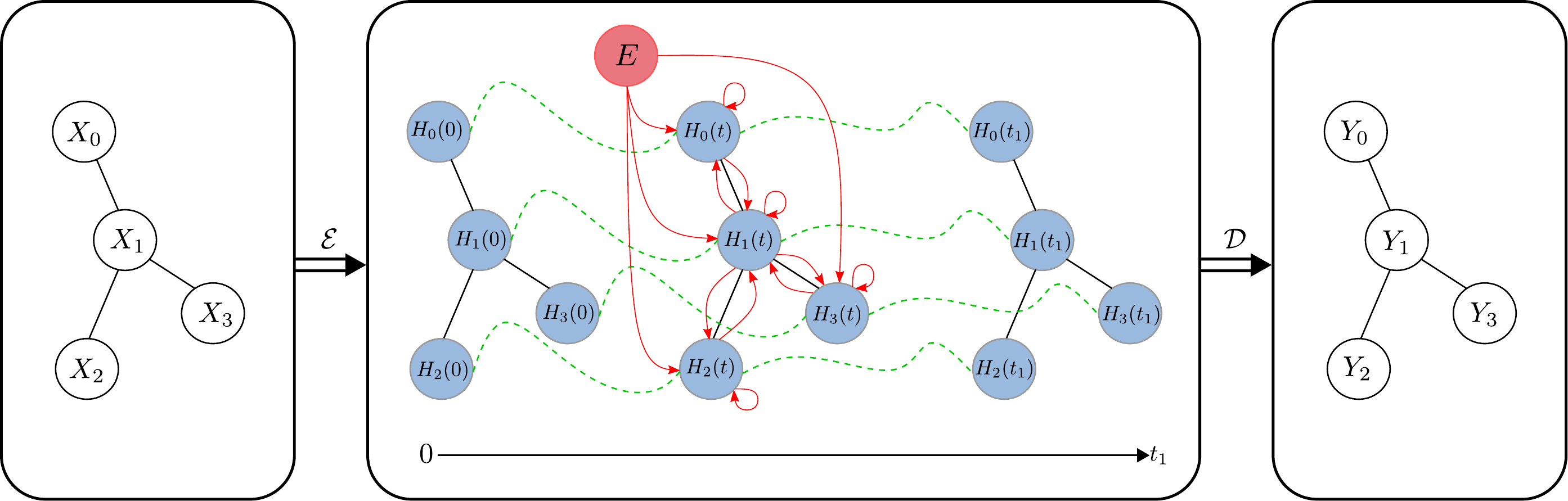}
    \caption{Architecture overview: The input to the model is a graph with node features, we initially encode these node features using a single neural network layer and ignoring the graph structure. Then we use a differential equation to change the representation over time, before projecting the representation using another single neural network layer and a softmax function to a one-hot encoding of the classes. The red lines represent the information transfer as defined by the ODE.}
    \label{fig:overview}
\end{figure*}

In this section, we introduce our proposed approach. The goal is to learn informative node representations $\mat{H} \in \mathbb{R}^{|V| \times d}$ for node classification. We first employ a neural encoder to project each node into a latent space based on its features, i.e. $\mat{E} = \mathcal{E}(\mat{X})$. Afterwards, $\mat{E}$ is treated as the initial value $\mat{H}(0)$, and an ODE is designed to define the continuous dynamics on node representations, where the long-term dependency between nodes can be effectively modelled. Finally, the node representations obtained at ending time $t_1$ (i.e. $\mat{H}(t_1)$) can be used for downstream applications through a decoder $\mathcal{D}$. The overall model architecture is summarized in Fig.~\ref{fig:overview}.

The key step of the framework is to design an effective ODE for defining the continuous dynamics on node representations, and thereby modelling the dependency of nodes. We design two such ODEs of increasing model capacity based on the intuition from existing diffusion-based methods on graphs. In the first ODE, each feature channel (i.e. dimension) of node representations evolves independently (see Sec.~\ref{sec: ODE without w}), whereas in the second ODE we also model the interaction of different feature channels (see Sec.~\ref{sec: ODE with w}).

\subsection{Case 1: Independent Feature Channels}
\label{sec: ODE without w}

Since different nodes in a graph are interconnected, a desirable ODE should take the graph structure into consideration, and allow information to propagate among different nodes. Motivated by existing diffusion-based methods on graphs (e.g.\ PageRank~\cite{page1999pagerank} and label propagation~\cite{zhou2004learning}), an effective way for characterizing the propagation process is to use the following step-wise propagation equations:
\begin{equation}
  \label{eq:starting point}
  \mat{H}_{n+1}= \mat{A} \mat{H}_{n}+\mat{H}_0,
\end{equation}
where $\mat{H}_n \in \mathbb{R}^{|V| \times d}$ is the embedding matrix for all the nodes at step $n$, and $\mat{H}_0=\mat{E}=\mathcal{E}(\mat{X})$ is the embedding matrix computed by the encoder $\mathcal{E}$.
Intuitively, each node at stage $n+1$ learns the node information from its neighbours through $\mat{A}\mat{H}_n$ and remembers its original node features through $\mat{H}_0$. This allows us to learn the graph structure without forgetting the original node features. The explicit formula of Eq.~\ref{eq:starting point} can be derived as follows:
\begin{equation}
  \label{eq:orig z(t)}
  \mat{H}_n = \left(\sum_{i=0}^n \mat{A}^i\right)\mat{H}_0 = (\mat{A} - \mat{I})^{-1}(\mat{A}^{n+1} - \mat{I})\mat{H}_0,
\end{equation}
where we see that the representation $\mat{H}_n$ at step $n$ incorporates all the information propagated up to $n$ steps with the initial representation $\mat{H}_0$.

Because of the effectiveness of the discrete propagation process in Eq.~\eqref{eq:orig z(t)}, we aim to extend the process to continuous cases by replacing $n$ with a continuous variable $t\in\mathbb{R}^+_0$, and further use an ODE to characterise such a continuous propagation dynamic. Intuitively, we view the summation in Eq.~\eqref{eq:orig z(t)} as a Riemann sum of an integral from time $0$ to time $t=n$, which allows us to naturally move from the discrete propagation process to the continuous case, as stated in the following proposition.

\begin{proposition}
\label{prop:no weights}
    The discrete dynamic in Eq.~\eqref{eq:orig z(t)}, where $\mat{A}$ has an eigenvalue decomposition, is a discretisation of the following ODE:
    \begin{equation}
        \label{eq:1st}
        \frac{\der \mat{H}(t)}{\der t} = \ln \mat{A} \mat{H}(t) + \mat{E},
    \end{equation}
    with the initial value $\mat{H}(0) = (\ln\mat{A})^{-1}(\mat{A}-\mat{I})\mat{E}$, where $\mat{E} = \mathcal{E}(\mat{X})$ is the output of the encoder $\mathcal{E}$.
\end{proposition}
We provide the proof in the Supplementary material. In practice, $\ln \mat{A}$ in Eq.~\eqref{eq:1st} is intractable to compute, hence we approximate it using the first order of the Taylor expansion, i.e.\ $\ln \mat{A}\approx (\mat{A}-\mat{I})$, which gives us:
\begin{equation}
    \label{eq:approx}
    \frac{\der \mat{H}(t)}{\der t} = (\mat{A}-\mat{I}) \mat{H}(t) + \mat{E},
\end{equation}
with the initial value being $\mat{H}(0) =  \mat{E}$. This is the ODE we use in our model \textbf{CGNN}. The intuition behind the ODE defined in Eq.~\ref{eq:approx} can be understood from an epidemic modelling perspective. The epidemic model aims at studying the dynamics of infection in a population. Typically, the model assumes that the infection of people is affected by three factors, i.e.\ the infection from neighbours, the natural recovery, and the natural physique of people. Suppose that we treat the latent vectors $\mat{H}(t)$ as the infection conditions of a group of people at time $t$, then the three factors can be naturally captured by three terms: $\mat{A}\mat{H}(t)$ for the infection from neighbours, $-\mat{H}(t)$ for natural recovery, and $\mat{E}$ for the natural physique. Therefore, the infection dynamics in a population can be intuitively modelled by our first-order ODE, in Eq.~\eqref{eq:1st}, indicating that the intuition of our ODE agrees with the epidemic model.

The ODE we use can be understood theoretically. Specifically, the node representation matrix $\mat{H}(t)$ at time $t$ has an analytical form, which is formally stated in the following proposition.

\begin{proposition}
    The analytical solution of the ODE defined in Eq.~\eqref{eq:approx} is given by:
    \begin{equation}
        \mat{H}(t)= (\mat{A}-\mat{I})^{-1}(e^{(\mat{A}-\mat{I})t}-\mat{I})\mat{E}+e^{(\mat{A}-\mat{I})t}\mat{E}
    \end{equation}
\end{proposition}
We prove the proposition in the Supplementary material. From the proposition, since the eigenvalues of $\mat{A}-\mat{I}$ are in the interval $[-1,0)$, as we increase $t$ to $\infty$, the exponential term $e^{(\mat{A}-\mat{I})t}$ will approach $\mat{0}$, i.e.\ $\lim_{t \to \infty } e^{(\mat{A}-\mat{I})t} = \mat{0}$. Therefore, for large
enough $t$ we can approximate $\mat{H}(t)$ as:
\begin{equation}
    \label{eq:analytical soln}
  \mat{H}(t) \approx (\mat{I}-\mat{A})^{-1}\mat{E} = \left(\sum_{i=0}^\infty \mat{A}^i \right)\mat{E}.
\end{equation}
Thus, $\mat{H}(t)$ can be seen as the summation of all different orders of propagated information (i.e.\ $\{\mat{A}^i \mat{E}\}_{i=1}^\infty$). In this way, our approach essentially has an infinite number of discrete propagation layers, allowing us to model global dependencies of nodes more effectively than existing GNNs. 

\textbf{Implementation}. In the node classification task, our decoder $\mathcal{D}$ to compute the node-label matrix $\mat{Y} = \mathcal{D}(\mat{H}(t_1))$ is a softmax classifier with the ReLU activation function \cite{nair2010rectified}.

Note the parameter $\alpha$ in Eq.~\eqref{eq: def A} decides the eigenvalues of $\mat{A}$, which thereby determines how quickly the higher order powers of $\mat{A}$ go to $0$, this also means that by specifying $\alpha$ per node we can control how much of the neighbourhood each node gets to see as smaller values $\alpha$ imply that the powers of $\mat{A}$ vanish faster. In our final model we learn these parameters $\alpha$.

\subsection{Case 2: Modelling the Interaction of Feature Channels}
\label{sec: ODE with w}

The ODE so far models different feature channels (i.e.\ dimensions of hidden representations) independently, where different channels are not able to interact with each other, and thus the ODE may not capture the correct dynamics of the graph. To allow the interaction between different feature channels, we are inspired by the success of a linear variant of GCN (i.e.\ Simple GCN \cite{wu2019}) and consider a more powerful discrete dynamic:
\begin{equation}
    \label{eq:starting point w}
    \mat{H}_{n + 1} = \mat{A}\mat{H}_n\mat{W} + \mat{H}_0,
\end{equation}
where $\mat{W} \in \mathbb{R}^{d \times d}$ is a weight matrix. Essentially, with $\mat{W}$, we are able to model the interactions of different feature channels during propagation, which increases the model capacity and allows us to learn more effectively representations of nodes.

Similar to the previous section, we extend the discrete propagation process in Eq.~\eqref{eq:starting point w} to continuous cases by viewing each $\mat{H}_n$ as a Riemann sum of an integral from time $0$ to time $t=n$, which yields the following proposition:
\begin{proposition}
    Suppose that the eigenvalue decompositions of $\mat{A},\mat{W}$ are $\mat{A} = \mat{P} \mat{\Lambda} \mat{P}^{-1} $ and $\mat{W} = \mat{Q} \mat{\Phi} \mat{Q}^{-1}$, respectively, then the discrete dynamic in Eq.~\eqref{eq:starting point w} is a discretisation of the following ODE:
    \begin{equation}
        \label{eq:ode w}
        \frac{\der \mat{H}(t)}{\der t} = \ln \mat{A} \mat{H}(t) + \mat{H}(t) \ln \mat{W} + \mat{E},
    \end{equation}
    where $\mat{E} = \mathcal{E}(\mat{X})$ is the output of the encoder $\mathcal{E}$ and with the initial value $\mat{H}(0) = \mat{P}\mat{F}\mat{Q}^{-1}$, where 
    \begin{equation}
        F_{ij} = \frac{\Lambda_{ii}\widetilde{E}_{ij}\Phi_{jj}-\widetilde{E}_{ij}}{\ln\Lambda_{ii}\Phi_{jj}},
    \end{equation}
    where $\widetilde{\mat{E}}=\mat{P}^{-1}\mat{E}\mat{Q}$.
\end{proposition}
The proof is provided in the Supplementary material. By making a first-order Taylor approximation to get rid of the matrix logarithm, we further obtain:
    \begin{equation}
        \label{eq:weight approx}
        \frac{\der \mat{H}(t)}{\der t} = (\mat{A}-\mat{I}) \mat{H}(t) + \mat{H}(t) (\mat{W}-\mat{I}) + \mat{E},
    \end{equation}
    with the initial value being $\mat{H}(0) = \mat{E}$. This is the ODE we use in our model \textbf{CGNN} with weights.
The ODEs of the form as in Eq.~\eqref{eq:weight approx} have been studied in some detail in the control theory literature, where they are known as the Sylvester differential equation~\cite{locatelli2001optimal,behr19}. Intuitively, $\mat{E}$ here would be the input into the system with the goal to get the system $\mat{H}$ into a desired state $\mat{H}(t)$. The matrices $\mat{A} - \mat{I}$ and $\mat{W} - \mat{I}$ describe the natural evolution of the system.

The ODE in Eq.~\eqref{eq:weight approx} also has appealing theoretical properties. Specifically, $\mat{H}(t)$ has an analytical form as shown in the following proposition.
\begin{proposition}
    Suppose that the eigenvalue decompositions of $\mat{A}-\mat{I},\mat{W}-\mat{I}$ are $\mat{A}-\mat{I} = \mat{P} \mat{\Lambda}' \mat{P}^{-1} $ and $\mat{W}-\mat{I} = \mat{Q} \mat{\Phi}' \mat{Q}^{-1}$, respectively, then the analytical soluton of the ODE in Eq.~\eqref{eq:weight approx} is given by:
    \begin{equation}
        \label{eq:weight soln final}
        \mat{H}(t) = e^{(\mat{A}-\mat{I})t}\mat{E}e^{(\mat{W}-\mat{I})t}+\mat{P}\mat{F}(t)\mat{Q}^{-1},
    \end{equation}
    where $\mat{F}(t) \in \mathbb{R}^{|V| \times d}$ with each element defined as follows:
    \begin{equation}
        F_{ij}(t) = \frac{\widetilde{E}_{ij}}{\Lambda_{ii}'+\Phi_{jj}'}e^{t({\Lambda}_{ii}'+\Phi_{jj}')}-\frac{\widetilde{{E}}_{ij}}{{\Lambda}_{ii}'+{\Phi}_{jj}'}
    \end{equation}
    where $\widetilde{\mat{E}}=\mat{P}^{-1}\mat{E}\mat{Q}$.
\end{proposition}

We prove the proposition in the Supplementary material. According to the definition of $\mat{A}$ and also our assumption about $\mat{W}$, the eigenvalues of $\mat{A} - \mat{I}$ and $\mat{W} - \mat{I}$ are in $(-1,0)$, and therefore $\Lambda_{i,i}' < 0$ for every $i$ and $\Phi_{j,j}' < 0$ for every $j$. Hence, as we increase $t$ to $\infty$, the exponential terms will approach 0, and hence for large enough $t$ we can approximate $\mat{H}(t)$ as: 
\begin{equation}
    \left(\mat{P}^{-1}\mat{H}(t)\mat{Q}\right)_{ij} \approx -\frac{\widetilde{{E}}_{ij}}{{\Lambda}_{ii}'+{\Phi}_{jj}'}.
    \label{eq:ode w used}
\end{equation}

Based on the above results, if $\mat{W}=\mat{I}$, then $\mat{H}(t)$ will converge to the same result as in Eq.~\eqref{eq:analytical soln}, and hence the ODE defined in Eq.~\eqref{eq:1st} is a special case of the ODE in Eq.~\eqref{eq:ode w}.

\textbf{Implementation}. We use the same decoder as for the case when $W=I$.

In practice, to enforce $\mat{W}$ to be a diagonalisable matrix with all the eigenvalues less than 1, we parameterise $\mat{W}$ as $\mat{W} = \mat{U} \text{diag}(\mat{M}) \mat{U}^T$, where $\mat{U} \in \mathbb{R}^{d \times d}$ is a learnable orthogonal matrix and $\mat{M} \in \mathbb{R}^d$ is a learnable vector, characterising the eigenvalues of $\mat{W}$. During training, we clamp the values of $\mat{M}$ to guarantee they are between $(0,1)$. To ensure $\mat{U}$ to be an orthogonal matrix, we follow previous work in \cite{cisse2017parseval, conn17} and perform the following secondary update of $\mat{U}$ after each main update during training:
\begin{equation}
    \mat{U} \leftarrow (1 + \beta) \mat{U} - \beta (\mat{U} \mat{U}^T) \mat{U},
\end{equation}
where $\beta$ is a hyperparameter, and the above secondary update enables $\mat{U}$ to be close to the manifold of orthogonal matrices after each training step.

Finally, to help stabilise training we use the idea from \citep{dupo2019} and add auxiliary dimensions to hidden representation only during the continuous propagation process. Specifically, we double the latent representation initialising the second half of the initial representation with 0 and throwing the result away after solving the continuous ODE. This very slightly improves results, but importantly stabilises training significantly (see Supplementary material).

\begin{table*}[t]
	\begin{center}
	\caption{Statistics of datasets.}
	\label{tab:dataset}
	\scalebox{1}
	{
		\begin{tabular}{c c c c c c c}\hline
		    \textbf{Dataset} & \textbf{\# Nodes} & \textbf{\# Edges} & \textbf{\# Features} & \textbf{\# Classes} & \textbf{Label Rate}  \\
	        \hline
	        Cora & 2,708 & 5,429 & 1,433 & 7 & 0.036\\
	        Citeseer & 3,327 & 4,732 & 3,703 & 6 & 0.052 \\
	        Pubmed & 19,717 & 44,338 & 500 & 3 & 0.003\\
	        NELL & 65,755 & 266,144 & 5,414 & 210 & 0.001\\
	        \hline
	    \end{tabular}
	}
	\end{center}
\end{table*}

\section{Discussion}

Our continuous model for information propagation has several advantages over previous discrete GNNs such as GCN:
\begin{enumerate}
    \item Robustness with time to over-smoothing;
    \item Learning global dependencies in the graph;
    \item $\alpha$ represents the ``diffusion'' constant, which is learned;
    \item Weights entangle channels continuously over time;
    \item Insight into the role of the restart distribution $\mat{H}_0$.
\end{enumerate}

\textbf{1.\ Robustness with time to over-smoothing}: Despite the effectiveness of the discrete propagation process, it has been shown in~\cite{liqi18} that the usage of discrete GCN layers can be tricky as the number $n$ of layers (time in the continuous case) is a critical choice. Theoretical work in \cite{oono2020} further showed that on dense graphs as the number of GCN layers grow there is exponential information loss in the node representations. In contrast, our method is experimentally not very sensitive to the integration time chosen and theoretically does not suffer from information loss as time goes to infinity.

\textbf{2.\ Global dependencies}: Recent work \cite{klic19,dehm19, xu2018} has shown that to improve on GNN it is necessary to build deeper networks to be able to learn long-range dependencies between nodes. Our work, thanks to the stability with time is able to learn global dependencies between nodes in the graph. Eq.~\eqref{eq:analytical soln} demonstrates that we propagate the information from all powers of the adjacency matrix, thus we are able to learn global dependencies.

\textbf{3.\ Diffusion constant}: The parameter $\alpha$ scales the matrix $\mat{A}$ (see Eq.~\eqref{eq: def A}), i.e.\ it controls the rate of diffusion. Hence, $\alpha$ controls the rate at which higher-order powers of $\mat{A}$ vanish. Since each node has its own parameter $\alpha$ that is learned, our model is able to control the diffusion, i.e.\ the weight of higher-order powers, for each node independently.

\textbf{4.\ Entangling channels during graph propagation}: The ODE with weights (Eq.~\eqref{eq:ode w}) allows the model to entangle the information from different channels over time. In addition, we are able to explain how the eigenvalues of the weight matrix affect the learned representation (Eq.~\eqref{eq:weight soln final}).

\textbf{5.\ Insight into the role of the restart distribution $\mat{H}_0$}: In both of our ODEs, Eq.~\eqref{eq:approx} and~\eqref{eq:weight approx}, the derivative depends on $\mat{E}$, which equals to the initial value $\mat{H}(0)$. To intuitively understand the effect of the initial value in our ODEs, consider an ODE without $\mat{E}$, i.e.\ $\mat{H}'(t) = (\mat{A}-I)\mat{H}(t)$. The analytical solution to the ODE $\mat{H}'(t) = (\mat{A}-I)\mat{H}(t)$ is given by $\mat{H}(t) = \exp[(\mat{A}-\mat{I})t]\mat{H}(0)$. Remembering that $\mat{A}-\mat{I}$ is simply a first order approximation of $\ln\mat{A}$, we can see that the analytical solution we are trying to approximate is $\mat{H}(t) = \mat{A}^t\mat{H}(0)$. Thus, the end time of the ODE now determines, which power of the $\mat{Adj}$ we learn. Indeed in our experiments we show that removing the term $\mat{H}(0)$ causes us to become very sensitive to the end time chosen rather than just needing a sufficiently large value (see Fig.~\ref{fig::time}).

\begin{table*}[t]
    \centering
    \caption{Node classification results on citation networks. $\star$ The values are taken from the original paper. $\star\star$ There is no standard test-validation-training split on this dataset, hence we generated a random one and used it across all experiments. $\star\star\star$ GAT ran out of memory on NELL.}
    \label{tab:node_class_res}
    \begin{tabular}{c|cccc}
         \hline\hline
         Model & Cora & Citeseer & Pubmed & NELL$^{\star\star}$\\\hline
         GAT-GODE$^{\star}$ & $83.3\pm0.3$ & $72.1\pm0.6$ & $79.1\pm0.5$ & -\\
         GCN-GODE$^{\star}$ & $81.8\pm0.3$ & $72.4\pm0.8$ & $80.1\pm0.3$ & -\\
         GCN & $81.8 \pm 0.8$ & $70.8 \pm 0.8$ & $80.0 \pm 0.5$ & $57.4 \pm 0.7$\\
         GAT$^{\star\star\star}$ & $82.6 \pm 0.7$ & $71.5 \pm 0.8$ & $77.8 \pm 0.6$ & -\\
         \hline
         CGNN discrete & $81.8\pm0.6$ & $70.0 \pm 0.5$ & $81.0\pm0.4$ & $50.9 \pm 3.9$\\
         \hline
         CGNN& $\boldsymbol{84.2\pm 1.0}$ & $72.6\pm 0.6$&$\boldsymbol{82.5\pm 0.4}$ & $65.4\pm 1.0$\\
         CGNN with weight & $83.9\pm0.7$ & $\boldsymbol{72.9\pm 0.6}$ & $82.1\pm0.5$ & $\boldsymbol{65.6\pm 0.9}$\\
         \hline\hline
    \end{tabular}
\end{table*}
\begin{table*}[t]
    \centering
    \caption{Node classification results on citation networks over 15 random splits. $\star$ GAT ran out of memory on NELL.}
    \label{tab:node_class_res_rand}
    \begin{tabular}{c|cccc}
         \hline\hline
         Model & Cora & Citeseer & Pubmed & NELL\\\hline
         GCN & $80.9\pm1.1$ & $72.0\pm 1.1$ & $81.5\pm1.4$ & $64.8 \pm 1.8$ \\
         GAT$^{\star}$ & $78.2\pm1.4$ & $71.9\pm1.4$ & $79.8\pm1.6$ & -\\
         \hline
         CGNN discrete & $81.7\pm1.5$ & $70.3\pm1.6$ & $81.8\pm1.7$ & $69.5\pm1.4$ \\
         \hline
         CGNN & $\boldsymbol{82.7\pm1.2}$ & $72.7\pm0.9$ & $\boldsymbol{83.2\pm1.4}$ & $\boldsymbol{73.4\pm1.0}$\\
         CGNN with weight & $82.1\pm1.3$ & $\boldsymbol{72.9\pm 0.9}$ & $82.7\pm1.4$ & $73.1\pm 0.9$\\
         \hline\hline
    \end{tabular}
\end{table*}

\section{Experiment}

In this section, we evaluate the performance of our proposed approach on the semi-supervised node classification task.

\subsection{Datasets and Experiment Settings}

In our experiment, we use four benchmark datasets for evaluation, including Cora, Citeseer, Pubmed, and NELL. Following existing studies~\cite{yang2016revisiting,kipf16,veli17}, we use the standard data splits from~\cite{yang2016revisiting} for Cora, Citeseer and Pubmed, where 20 nodes of each class are used for training and another 500 labeled nodes are used for validation. For the NELL dataset, as the data split used in~\cite{yang2016revisiting} is not available, we create a new split for experiment. The results are in Table~\ref{tab:node_class_res}. We further run experiments with random splits on the same datasets in Table~\ref{tab:node_class_res_rand}. The statistics of the datasets are summarized in Table~\ref{tab:dataset}. Accuracy is used as the evaluation metric.

\subsection{Compared Algorithms}

\smallskip
\noindent \textbf{Discrete GNNs}: For standard graph neural networks which model the discrete dynamic of node representations, we mainly compare with the Graph Convolutional Network (\textbf{GCN})~\cite{kipf16} and the Graph Attention Network (\textbf{GAT})~\cite{veli17}, which are the most representative methods.

\smallskip
\noindent \textbf{Continuous GNNs}: There is also a recent concurrent work~\cite{zhua2020} which learns node representations through modelling the continuous dynamics of node representations, where the ODE is parameterised by a graph neural network. We also compare with this method (\textbf{GODE}).

\smallskip
\noindent \textbf{CGNN}: For our proposed Continuous Graph Neural Network (CGNN), we consider a few variants. Specifically, \textbf{CGNN} leverages the ODE in Eq.~\eqref{eq:approx} to define the continuous dynamic of node representations, where different feature channels are independent. \textbf{CGNN with weight} uses the ODE in Eq.~\eqref{eq:weight approx}, which allows different feature channels to interact with each other. 
We also compare with \textbf{CGNN discrete}, which uses the discrete propagation process defined in Eq.~\eqref{eq:starting point} for node representation learning (with $n=50$).

\subsection{Parameter Settings}

We do a random hyperparameter search using the \textsc{Orion}~\cite{orion, lili18} framework with 50 retries. The mean accuracy over 10 runs is reported for each dataset in Table~\ref{tab:node_class_res}.

\subsection{Results}

\smallskip
\noindent \textbf{1. Comparison with existing methods.} The main results of different compared algorithms are summarized in Table~\ref{tab:node_class_res}. Compared with standard discrete graph neural networks, such as GCN and GAT, our approach achieves significantly better results in most cases. The reason is that our approach can better capture the long-term dependency of different nodes. Besides, our approach also outperforms the concurrent work GODE on Cora and Pubmed. This is because our ODEs are designed based on our prior knowledge about information propagation in graphs, whereas GODE parameterises the ODE by straightforwardly using an existing graph neural network (e.g.\ GCN or GAT), which may not effectively learn to propagate information in graphs. Overall, our approach achieves comparable results to state-of-the-art graph neural networks on node classification.

\begin{figure}[t]
	\centering
	\subfigure[Cora]{
		\label{fig:time-cora}
		\includegraphics[width=0.4\textwidth]{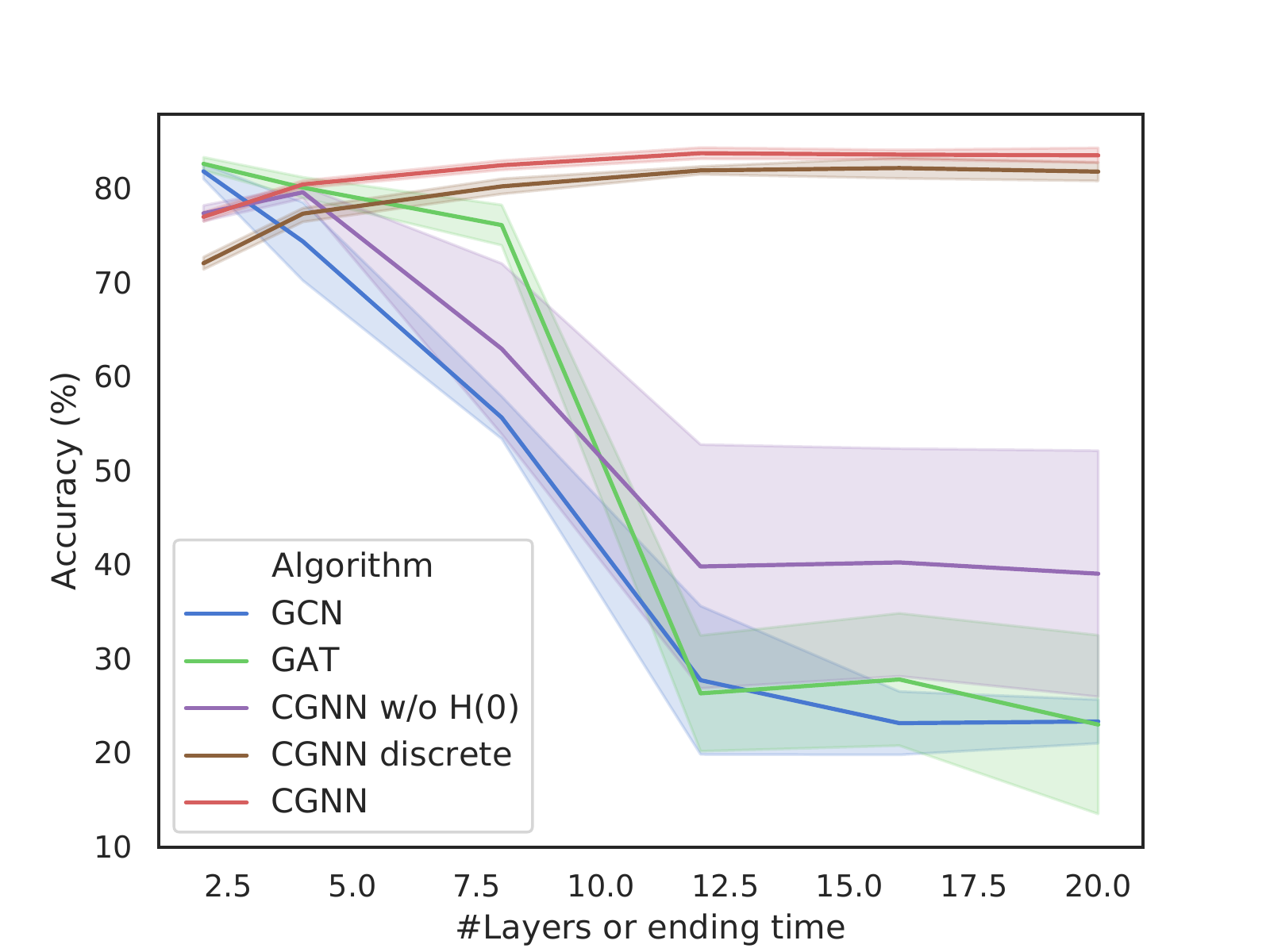}
	}\\
	\vspace{-12pt}
	\subfigure[Pubmed]{
		\label{fig::time-pubmed}
		\includegraphics[width=0.4\textwidth]{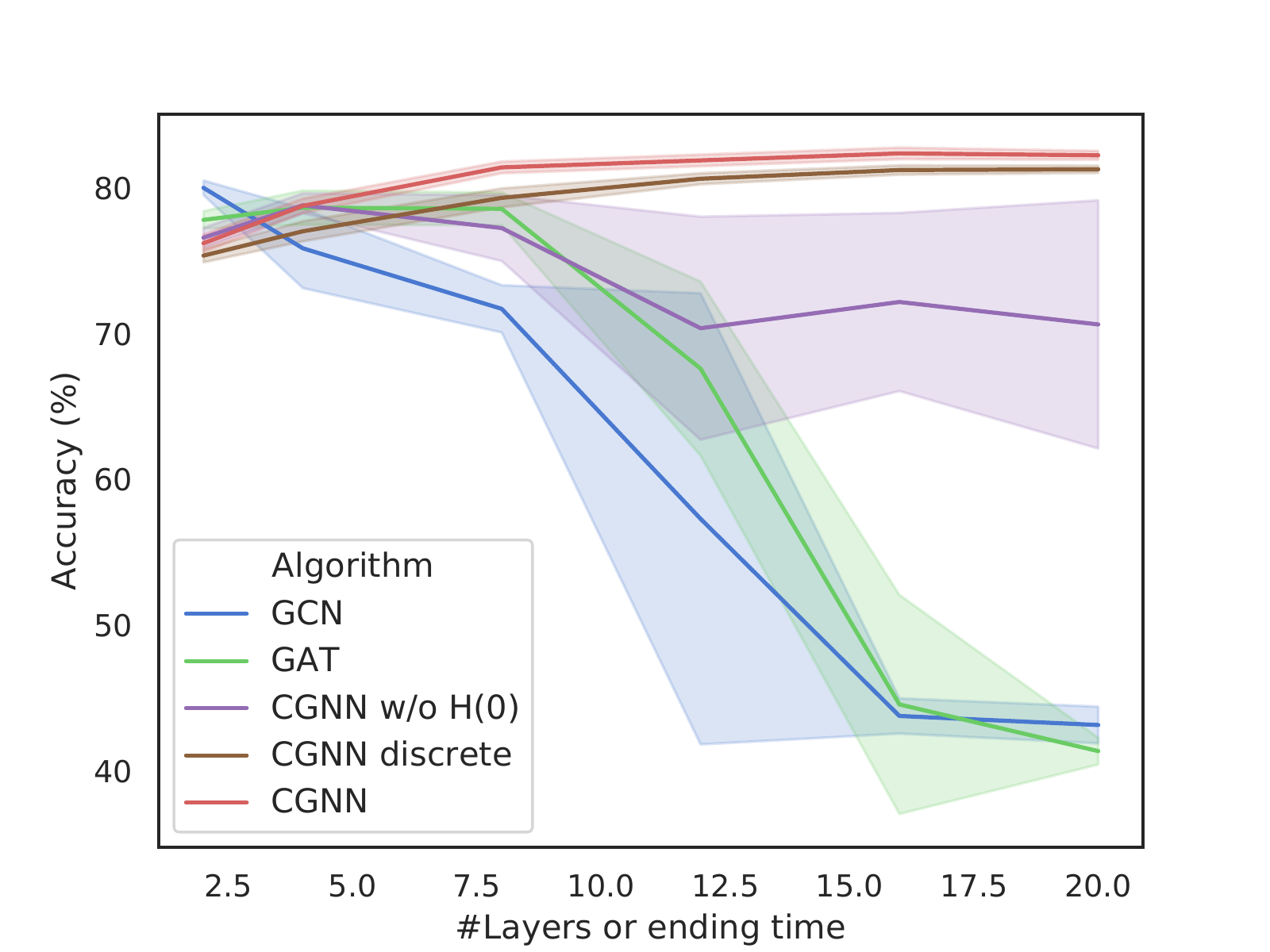}
	}
	\caption{Performance w.r.t. \#layers or ending time. Note that the red line also has error bars, but they are very small.}
	\label{fig::time}
	\vspace{-0.5cm}
\end{figure}
\begin{figure}[t]
	\centering
	\subfigure[Cora]{
		\label{fig:mem-cora}
		\includegraphics[width=0.4\textwidth]{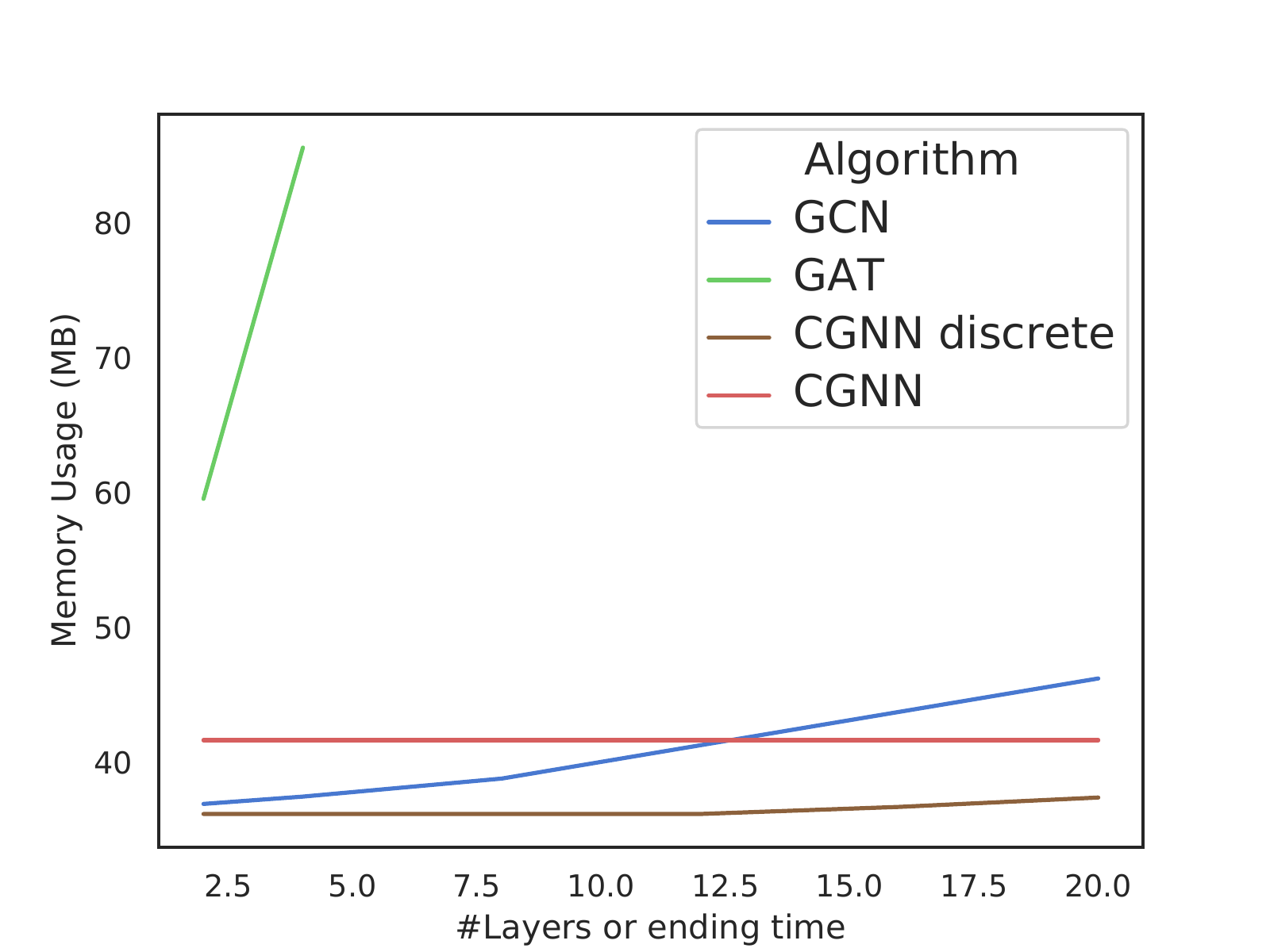}
	}\\
	\vspace{-12pt}
	\subfigure[Pubmed]{
		\label{fig::mem-pubmed}
		\includegraphics[width=0.4\textwidth]{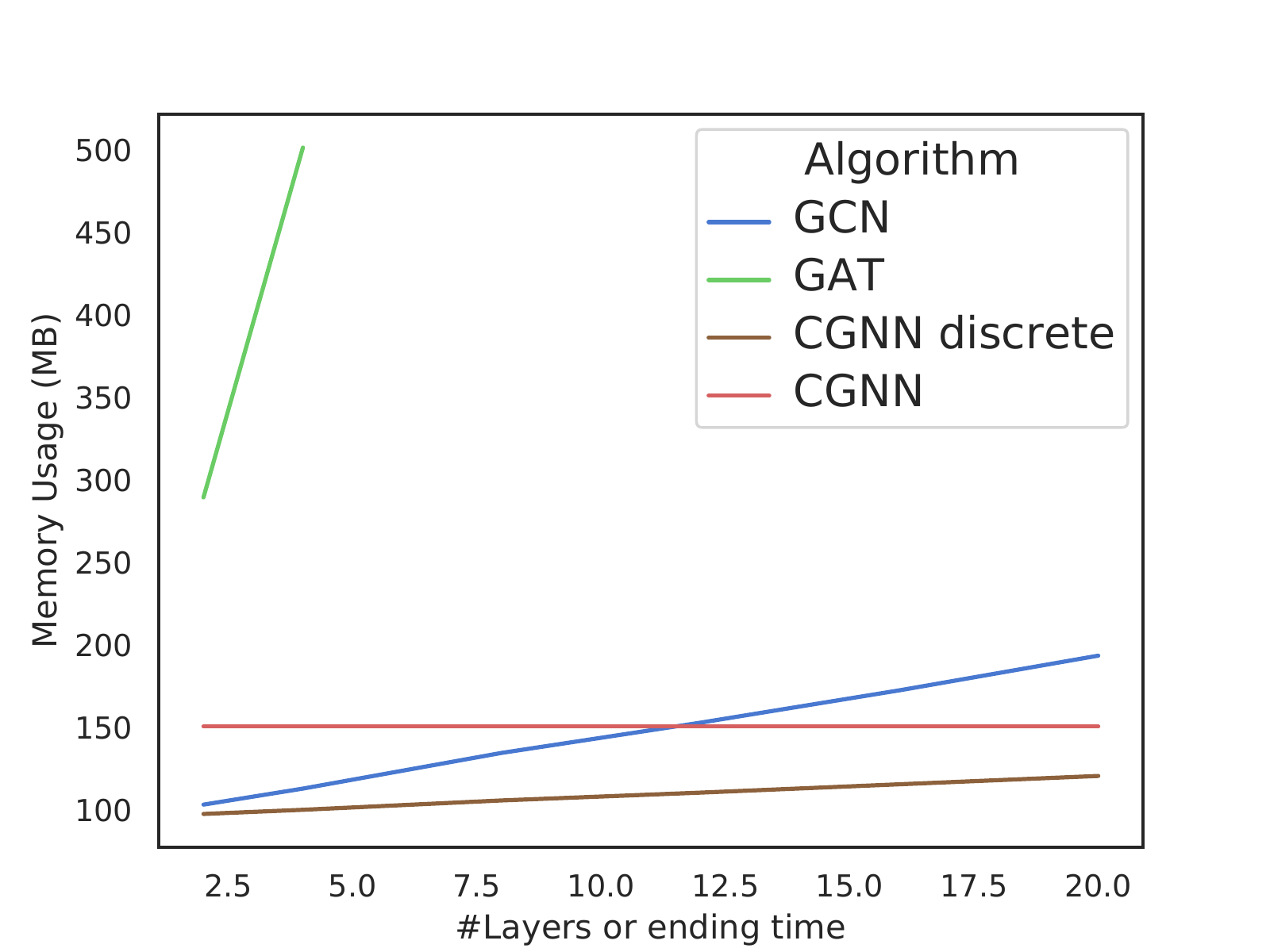}
	}
	\caption{Memory usage w.r.t.\ \#layers or ending time.}
	\label{fig::mem}
\end{figure}
\smallskip
\noindent \textbf{2. Comparison of CGNN and its variants.}
The ODEs in CGNN are inspired by the discrete propagation process in Eq.~\eqref{eq:starting point}, which can be directly used for modelling the dynamic on node representations. Compared with this variant (CGNN discrete), CGNN achieves much better results on all the datasets, showing that modelling the dynamic on nodes continuously is more effective for node representation learning. Furthermore, comparing the ODEs with or without modelling the interactions of feature channels (CGNN with weight and CGNN respectively), we see that their results are close. A possible reason is that the datasets used in experiments are quite easy, and thus modelling the interactions of feature channels (CGNN with weight) does not bring much gain in performance. We anticipate CGNN with weight could be more effective on more challenging graphs, and we leave it as future work to verify this.

\smallskip
\noindent \textbf{3. Performance with respect to time steps.} One major advantage of CGNN over existing methods is that it is robust to the over-smoothing problem. Next, we systematically justify this point by presenting the performance of different methods under different numbers of layers (e.g.\ GCN and GAT) or the ending time $t$ (e.g.\ CGNN and its variants). 

The results on Cora and Pubmed are presented in Fig.~\ref{fig::time}. For GCN and GAT, the optimal results are achieved when the number of layers is 2 or 3. If we stack more layers, the results drop significantly due to the over-smoothing problem. Therefore, GCN and GAT are only able to leverage information within 3 steps for each node to learn the representation. In contrast to them, the performance of CGNN is more stable and the optimal results are achieved when $t>10$, which shows that CGNN is robust to over-smoothing and can effectively model long-term dependencies of nodes. To demonstrate this we use the ODE $\mat{H}'(t)=(\mat{A}-\mat{I})\mat{H}(0)$ with $\mat{H}(0) =\mat{E}$, which gets much worse results (CGNN w/o $\mat{H}(0)$), showing the importance of the initial value for modelling the continuous dynamic. Finally, CGNN also outperforms the variant which directly models the discrete dynamic in Eq.~\eqref{eq:starting point} (CGNN discrete), which demonstrates the advantage of our continuous approach.

\smallskip
\noindent \textbf{4. Memory Efficiency.}
Finally, we compare the memory efficiency of different methods on Cora and Pubmed in Fig.~\ref{fig::mem}. For all the methods modelling the discrete dynamic, i.e.\ GCN, GAT, and CGNN discrete, the memory cost is linear to the number of discrete propagation layers. In contrast, through using the adjoint method~\cite{pontryagin2018mathematical} for optimization, CGNN has a constant memory cost and the cost is quite small, which is hence able to model long-term node dependency on large graphs.
\section{Conclusion}

In this paper, we build the connection between recent graph neural networks and traditional dynamic systems. Based on the connection, we further propose continuous graph neural networks (CGNNs), which generalise existing discrete graph neural networks to continuous cases through defining the evolution of node representations with ODEs. Our ODEs are motivated by existing diffusion-based methods on graphs, where two different ways are considered, including different feature channels change independently or interact with each other. Extensive theoretical and empirical analysis prove the effectiveness of CGNN over many existing methods. 
Our current approach assumes that connected nodes are similar (`homophily' assumption), we leave it for future work to be able to learn more complex non-linear relationships such as can be found in molecules~\cite{gilmer2017neural} or knowledge graphs~\cite{sun2019rotate}.

\subsubsection*{Acknowledgement}
This project is supported by the Natural Sciences and Engineering Research Council (NSERC) Discovery Grant, the Canada CIFAR AI Chair Program, collaboration grants between Microsoft Research and Mila, Amazon Faculty Research Award, Tencent AI Lab Rhino-Bird Gift Fund and a NRC Collaborative R\&D Project (AI4D-CORE-06). 

We also would like to thank Joey Bose, Alexander Tong, Emma Rocheteau, and
Andreea Deac for useful comments on the manuscript and S\'ekou-Oumar Kaba for pointing out a mistake in one equation.

\FloatBarrier

\bibliographystyle{plainnat}
\bibliography{ref}

\begin{thebibliography}{37}
\providecommand{\natexlab}[1]{#1}
\providecommand{\url}[1]{\texttt{#1}}
\expandafter\ifx\csname urlstyle\endcsname\relax
  \providecommand{\doi}[1]{doi: #1}\else
  \providecommand{\doi}{doi: \begingroup \urlstyle{rm}\Url}\fi

\bibitem[Avelar et~al.(2019)Avelar, Tavares, Gori, and Lamb]{avel2019}
Pedro H.~C. Avelar, Anderson~R. Tavares, Marco Gori, and Luis~C. Lamb.
\newblock Discrete and continuous deep residual learning over graphs, 2019.

\bibitem[Behr et~al.(2019)Behr, Benner, and Heiland]{behr19}
Maximilian Behr, Peter Benner, and Jan Heiland.
\newblock Solution formulas for differential sylvester and lyapunov equations.
\newblock \emph{Calcolo}, 56\penalty0 (4):\penalty0 51, 2019.

\bibitem[Bouthillier et~al.(2019)Bouthillier, Tsirigotis, Corneau-Tremblay,
  Delaunay, Noukhovitch, Askari, Henderson, Suhubdy, Bastien, and
  Lamblin]{orion}
Xavier Bouthillier, Christos Tsirigotis, Fran{\c{c}}ois Corneau-Tremblay,
  Pierre Delaunay, Michael Noukhovitch, Reyhane Askari, Peter Henderson, Dendi
  Suhubdy, Fr{\'e}d{\'e}ric Bastien, and Pascal Lamblin.
\newblock {Oríon - Asynchronous Distributed Hyperparameter Optimization}.
\newblock \url{https://github.com/Epistimio/orion}, September 2019.

\bibitem[Chen et~al.(2018)Chen, Rubanova, Bettencourt, and Duvenaud]{chen18}
Tian~Qi Chen, Yulia Rubanova, Jesse Bettencourt, and David~K Duvenaud.
\newblock Neural ordinary differential equations.
\newblock In S.~Bengio, H.~Wallach, H.~Larochelle, K.~Grauman, N.~Cesa-Bianchi,
  and R.~Garnett, editors, \emph{Advances in Neural Information Processing
  Systems 31}, pages 6571--6583. Curran Associates, Inc., 2018.
\newblock URL
  \url{http://papers.nips.cc/paper/7892-neural-ordinary-differential-equations.pdf}.

\bibitem[Chung and Graham(1997)]{chun97}
Fan~RK Chung and Fan~Chung Graham.
\newblock \emph{Spectral graph theory}.
\newblock Number~92. American Mathematical Soc., 1997.

\bibitem[Cisse et~al.(2017)Cisse, Bojanowski, Grave, Dauphin, and
  Usunier]{cisse2017parseval}
Moustapha Cisse, Piotr Bojanowski, Edouard Grave, Yann Dauphin, and Nicolas
  Usunier.
\newblock Parseval networks: Improving robustness to adversarial examples.
\newblock In \emph{Proceedings of the 34th International Conference on Machine
  Learning-Volume 70}, pages 854--863. JMLR. org, 2017.

\bibitem[Conneau et~al.(2017)Conneau, Lample, Ranzato, Denoyer, and
  J{\'e}gou]{conn17}
Alexis Conneau, Guillaume Lample, Marc'Aurelio Ranzato, Ludovic Denoyer, and
  Herv{\'e} J{\'e}gou.
\newblock Word translation without parallel data.
\newblock \emph{arXiv preprint arXiv:1710.04087}, 2017.

\bibitem[Defferrard et~al.(2016)Defferrard, Bresson, and Vandergheynst]{deff16}
Michaël Defferrard, Xavier Bresson, and Pierre Vandergheynst.
\newblock Convolutional neural networks on graphs with fast localized spectral
  filtering, 2016.

\bibitem[Dehmamy et~al.(2019)Dehmamy, Barab{\'a}si, and Yu]{dehm19}
Nima Dehmamy, Albert-L{\'a}szl{\'o} Barab{\'a}si, and Rose Yu.
\newblock Understanding the representation power of graph neural networks in
  learning graph topology.
\newblock In \emph{Advances in Neural Information Processing Systems}, pages
  15387--15397, 2019.

\bibitem[Deng et~al.(2019)Deng, Nawhal, Meng, and Mori]{deng19}
Zhiwei Deng, Megha Nawhal, Lili Meng, and Greg Mori.
\newblock Continuous graph flow for flexible density estimation.
\newblock \emph{arXiv preprint arXiv:1908.02436}, 2019.

\bibitem[Dupont et~al.(2019)Dupont, Doucet, and Teh]{dupo2019}
Emilien Dupont, Arnaud Doucet, and Yee~Whye Teh.
\newblock Augmented neural odes.
\newblock \emph{arXiv preprint arXiv:1904.01681}, 2019.

\bibitem[Gilmer et~al.(2017)Gilmer, Schoenholz, Riley, Vinyals, and
  Dahl]{gilmer2017neural}
Justin Gilmer, Samuel~S Schoenholz, Patrick~F Riley, Oriol Vinyals, and
  George~E Dahl.
\newblock Neural message passing for quantum chemistry.
\newblock In \emph{Proceedings of the 34th International Conference on Machine
  Learning-Volume 70}, pages 1263--1272. JMLR. org, 2017.

\bibitem[Kipf and Welling(2016)]{kipf16}
Thomas~N Kipf and Max Welling.
\newblock Semi-supervised classification with graph convolutional networks.
\newblock \emph{arXiv preprint arXiv:1609.02907}, 2016.

\bibitem[Klicpera et~al.(2019)Klicpera, Weißenberger, and Günnemann]{klic19}
Johannes Klicpera, Stefan Weißenberger, and Stephan Günnemann.
\newblock Diffusion improves graph learning, 2019.

\bibitem[Li et~al.(2018{\natexlab{a}})Li, Jamieson, Rostamizadeh, Gonina,
  Hardt, Recht, and Talwalkar]{lili18}
Liam Li, Kevin Jamieson, Afshin Rostamizadeh, Katya Gonina, Moritz Hardt,
  Benjamin Recht, and Ameet Talwalkar.
\newblock Massively parallel hyperparameter tuning, 2018{\natexlab{a}}.
\newblock URL \url{https://openreview.net/forum?id=S1Y7OOlRZ}.

\bibitem[Li et~al.(2018{\natexlab{b}})Li, Han, and Wu]{liqi18}
Qimai Li, Zhichao Han, and Xiao-Ming Wu.
\newblock Deeper insights into graph convolutional networks for semi-supervised
  learning.
\newblock In \emph{Thirty-Second AAAI Conference on Artificial Intelligence},
  2018{\natexlab{b}}.

\bibitem[Locatelli(2001)]{locatelli2001optimal}
Arturo Locatelli.
\newblock \emph{Optimal Control. An Introduction}.
\newblock Birkhauser Verlag, 2001.

\bibitem[Luan et~al.(2019)Luan, Zhao, Chang, and Precup]{luan19}
Sitao Luan, Mingde Zhao, Xiao-Wen Chang, and Doina Precup.
\newblock Break the ceiling: Stronger multi-scale deep graph convolutional
  networks.
\newblock In H.~Wallach, H.~Larochelle, A.~Beygelzimer, F.~d\textquotesingle
  Alch\'{e}-Buc, E.~Fox, and R.~Garnett, editors, \emph{Advances in Neural
  Information Processing Systems 32}, pages 10943--10953. Curran Associates,
  Inc., 2019.
\newblock URL
  \url{http://papers.nips.cc/paper/9276-break-the-ceiling-stronger-multi-scale-deep-graph-convolutional-networks.pdf}.

\bibitem[Marcheggiani and Titov(2017)]{marcheggiani2017encoding}
Diego Marcheggiani and Ivan Titov.
\newblock Encoding sentences with graph convolutional networks for semantic
  role labeling.
\newblock In \emph{Proceedings of the 2017 Conference on Empirical Methods in
  Natural Language Processing}, pages 1506--1515, 2017.

\bibitem[Nair and Hinton(2010)]{nair2010rectified}
Vinod Nair and Geoffrey~E Hinton.
\newblock Rectified linear units improve restricted boltzmann machines.
\newblock In \emph{Proceedings of the 27th international conference on machine
  learning (ICML-10)}, pages 807--814, 2010.

\bibitem[NT and Maehara(2019)]{nt2019}
Hoang NT and Takanori Maehara.
\newblock Revisiting graph neural networks: All we have is low-pass filters,
  2019.

\bibitem[Oono and Suzuki(2020)]{oono2020}
Kenta Oono and Taiji Suzuki.
\newblock Graph neural networks exponentially lose expressive power for node
  classification.
\newblock In \emph{International Conference on Learning Representations}, 2020.
\newblock URL \url{https://openreview.net/forum?id=S1ldO2EFPr}.

\bibitem[Page et~al.(1999)Page, Brin, Motwani, and Winograd]{page1999pagerank}
Lawrence Page, Sergey Brin, Rajeev Motwani, and Terry Winograd.
\newblock The pagerank citation ranking: Bringing order to the web.
\newblock Technical report, Stanford InfoLab, 1999.

\bibitem[Poli et~al.(2019)Poli, Massaroli, Park, Yamashita, Asama, and
  Park]{poli2019}
Michael Poli, Stefano Massaroli, Junyoung Park, Atsushi Yamashita, Hajime
  Asama, and Jinkyoo Park.
\newblock Graph neural ordinary differential equations, 2019.

\bibitem[Pontryagin(2018)]{pontryagin2018mathematical}
Lev~Semenovich Pontryagin.
\newblock \emph{Mathematical theory of optimal processes}.
\newblock Routledge, 2018.

\bibitem[Qu et~al.(2019)Qu, Bengio, and Tang]{qu2019}
Meng Qu, Yoshua Bengio, and Jian Tang.
\newblock Gmnn: Graph markov neural networks.
\newblock In \emph{ICML}, 2019.

\bibitem[Sun et~al.(2019)Sun, Deng, Nie, and Tang]{sun2019rotate}
Zhiqing Sun, Zhi-Hong Deng, Jian-Yun Nie, and Jian Tang.
\newblock Rotate: Knowledge graph embedding by relational rotation in complex
  space.
\newblock \emph{arXiv preprint arXiv:1902.10197}, 2019.

\bibitem[Veli{\v{c}}kovi{\'c} et~al.(2017)Veli{\v{c}}kovi{\'c}, Cucurull,
  Casanova, Romero, Lio, and Bengio]{veli17}
Petar Veli{\v{c}}kovi{\'c}, Guillem Cucurull, Arantxa Casanova, Adriana Romero,
  Pietro Lio, and Yoshua Bengio.
\newblock Graph attention networks.
\newblock \emph{arXiv preprint arXiv:1710.10903}, 2017.

\bibitem[Wijesinghe and Wang(2019)]{wije19}
W.~O. K. Asiri~Suranga Wijesinghe and Qing Wang.
\newblock Dfnets: Spectral cnns for graphs with feedback-looped filters.
\newblock In H.~Wallach, H.~Larochelle, A.~Beygelzimer, F.~d\textquotesingle
  Alch\'{e}-Buc, E.~Fox, and R.~Garnett, editors, \emph{Advances in Neural
  Information Processing Systems 32}, pages 6007--6018. Curran Associates,
  Inc., 2019.
\newblock URL
  \url{http://papers.nips.cc/paper/8834-dfnets-spectral-cnns-for-graphs-with-feedback-looped-filters.pdf}.

\bibitem[Wu et~al.(2019)Wu, Zhang, Souza~Jr, Fifty, Yu, and Weinberger]{wu2019}
Felix Wu, Tianyi Zhang, Amauri Holanda~de Souza~Jr, Christopher Fifty, Tao Yu,
  and Kilian~Q Weinberger.
\newblock Simplifying graph convolutional networks.
\newblock \emph{arXiv preprint arXiv:1902.07153}, 2019.

\bibitem[Xu et~al.(2018)Xu, Li, Tian, Sonobe, Kawarabayashi, and
  Jegelka]{xu2018}
Keyulu Xu, Chengtao Li, Yonglong Tian, Tomohiro Sonobe, Ken-ichi Kawarabayashi,
  and Stefanie Jegelka.
\newblock Representation learning on graphs with jumping knowledge networks.
\newblock \emph{arXiv preprint arXiv:1806.03536}, 2018.

\bibitem[Yang et~al.(2016)Yang, Cohen, and Salakhudinov]{yang2016revisiting}
Zhilin Yang, William Cohen, and Ruslan Salakhudinov.
\newblock Revisiting semi-supervised learning with graph embeddings.
\newblock In \emph{ICML}, 2016.

\bibitem[Yao et~al.(2019)Yao, Mao, and Luo]{yao2019graph}
Liang Yao, Chengsheng Mao, and Yuan Luo.
\newblock Graph convolutional networks for text classification.
\newblock In \emph{Proceedings of the AAAI Conference on Artificial
  Intelligence}, volume~33, pages 7370--7377, 2019.

\bibitem[Zhang and Chen(2018)]{zhang2018link}
Muhan Zhang and Yixin Chen.
\newblock Link prediction based on graph neural networks.
\newblock In \emph{Advances in Neural Information Processing Systems}, pages
  5165--5175, 2018.

\bibitem[Zhou et~al.(2004)Zhou, Bousquet, Lal, Weston, and
  Sch{\"o}lkopf]{zhou2004learning}
Dengyong Zhou, Olivier Bousquet, Thomas~N Lal, Jason Weston, and Bernhard
  Sch{\"o}lkopf.
\newblock Learning with local and global consistency.
\newblock In \emph{Advances in neural information processing systems}, pages
  321--328, 2004.

\bibitem[Zhou et~al.(2018)Zhou, Cui, Zhang, Yang, Liu, and Sun]{zhou2018graph}
Jie Zhou, Ganqu Cui, Zhengyan Zhang, Cheng Yang, Zhiyuan Liu, and Maosong Sun.
\newblock Graph neural networks: A review of methods and applications.
\newblock \emph{arXiv preprint arXiv:1812.08434}, 2018.

\bibitem[Zhuang et~al.(2020)Zhuang, Dvornek, Li, and Duncan]{zhua2020}
Juntang Zhuang, Nicha Dvornek, Xiaoxiao Li, and James~S. Duncan.
\newblock Ordinary differential equations on graph networks, 2020.
\newblock URL \url{https://openreview.net/forum?id=SJg9z6VFDr}.

\end{thebibliography}

\newpage
\onecolumn
\appendix

\section{Proof of Proposition 1 and 2}
\label{sec:no weights}
\textit{Proof of Proposition 1:}
The starting point is to see 
\begin{equation}
    \mat{H}_n = \left(\sum_{i=0}^n \mat{A}^i\right)\mat{H}_0
\end{equation}
as a Riemann sum, i.e.:
\begin{equation}
    \sum_{i=1}^{n+1}\mat{A}^{0+(i-1)\cdot \Delta t }\mat{E}\Delta t,
\end{equation}
where $\Delta t= \frac{t+1-0}{n+1}$ with $t=n$ and $\mat{E}=\mat{H}_0$ as before. So now letting $n\to\infty$ we get the following integral
\begin{equation}
    \mat{H}(t)=\int_0^{t+1}\mat{A}^s\mat{E}\mathrm ds,
\end{equation}
The derivative is then given by
\begin{equation}
    \frac{\der \mat{H}(t)}{\der t} = \mat{A}^{t+1}\mat{E}.
\end{equation}
However, $\mat{A}^{t+1}$ is intractable in practice to compute for non-integer $t$, hence we solve the ODE by considering the second order ODE and then integrating again.
\begin{equation}
\frac{\der^2 \mat{H}(t)}{\der t^2} = \ln\mat{A}\mat{A}^{t+1}\mat{E}=\ln\mat{A}\frac{\der \mat{H}(t)}{\der t}
\end{equation}
Now, integrating again
\begin{equation}
    \frac{\der \mat{H}(t)}{\der t} = \ln\mat{A}\mat{H}(t) +\textit{const}
\end{equation}
and solving for the constant using the fact that
\begin{equation}
    \mat{H}(0)=\int_0^1\mat{A}^s\mat{E}\mathrm ds=\frac{\mat{A}-\mat{I}}{\ln\mat{A}}\mat{E},
\end{equation}
we get that
\begin{equation}
    \left.\frac{\der \mat{H}(t)}{\der t}\right|_{t=0} = \mat{A}\mat{E}=\ln\mat{A}\mat{H}(0)+\textit{const} \implies \textit{const}=\mat{E}.
\end{equation}
Thus, we get the final ODE 
\begin{equation}
    \frac{\der \mat{H}(t)}{\der t} = \ln\mat{A}\mat{H}(t) +\mat{E}
\end{equation}
\phantom{QED}\hfill$\Box$

\textit{Proof of Proposition 2:}
To solve the ODE 
\begin{equation}
    \frac{\der \mat{H}(t)}{\der t} = (\mat{A}-\mat{I}) \mat{H}(t) + \mat{E}
\end{equation}
we will make use of a \emph{Ansatz} and use the integrating factor $\exp(-(\mat{A}-\mat{I})t)$.
\begin{align}
    e^{-(\mat{A}-\mat{I})t}\frac{\der \mat{H}(t)}{\der t} &= e^{-(\mat{A}-\mat{I})t}(\mat{A}-\mat{I})\mat{H}(t) +e^{-(\mat{A}-\mat{I})t}\mat{E}\\
    e^{-(\mat{A}-\mat{I})t}\frac{\der \mat{H}(t)}{\der t} - e^{-(\mat{A}-\mat{I})t}(\mat{A}-\mat{I})\mat{H}(t)&= e^{-(\mat{A}-\mat{I})t}\mat{E}\\
    e^{-(\mat{A}-\mat{I})t}\mat{H}(t)-\mat{E}&= -(\mat{A}-\mat{I})^{-1}(e^{-(\mat{A}-\mat{I})t}-I)\mat{E}\\
    \mat{H}(t)&= (\mat{A}-\mat{I})^{-1}(e^{(\mat{A}-\mat{I})t}-I)\mat{E}+e^{(\mat{A}-\mat{I})t}\mat{E}
\end{align}
\phantom{QED}\hfill$\Box$

\section{Proof of Proposition 3 and 4}
\label{sec:weights}
To prove \textit{Proposition} 3 and 4, we first prove the following Lemmata:
\begin{lemma}
\label{lemma:sylvester}
    The analytical solution of the following ODE, 
    \begin{equation}
    \label{eq:sylvester ode}
    \frac{\partial \mat{H}(t)}{\partial t} = \mat{B} \mat{H}(t) + \mat{H}(t) \mat{C} + \mat{D},
\end{equation}
    where $\mat{B}$ and $\mat{C}$ have eigenvalue decompositions $\mat{P}\mat{\Lambda}\mat{P}^{-1}$ and $\mat{Q}\mat{\Phi}\mat{Q}^{-1}$ respectively, with initial value $\mat{H}(0)$ is:
    \begin{equation}
        \mat{H}(t) =  e^{\mat{B}t}\mat{H}(0)e^{\mat{C}t}+ \mat{P}\mat{F}(t)\mat{Q}^{-1},
    \end{equation}
    where 
    \begin{equation}
        F_{ij}(t) = \frac{\widetilde{D}_{ij}}{\Lambda_{ii}+\Phi_{jj}}e^{t({\Lambda}_{ii}+\Phi_{jj})}-\frac{\widetilde{{D}}_{ij}}{{\Lambda}_{ii}+{\Phi}_{jj}}.
    \end{equation}
    with $\widetilde{\mat{D}}=\mat{P}^{-1}\mat{D}\mat{Q}$.
\end{lemma}
\textit{Proof:} First note that the ODE in Eq.~\eqref{eq:sylvester ode} is known as the Sylvester ODE \cite{behr19} with analytical solution:
\begin{equation}
    \mat{H}(t) = e^{\mat{B}t}\mat{H}(0)e^{\mat{C}t} + \int_0^t  e^{\mat{B}(t-s)}\mat{D}e^{\mat{C}(t-s)} \mathrm ds.
\end{equation}
Hence, to prove Lemma~\ref{lemma:sylvester} it remains to solve the integral using the help our assumptions.
\begin{align}
    \int_0^t  e^{\mat{B}(t-s)}\mat{D}e^{\mat{C}(t-s)} \mathrm ds &= \int_0^t \mat{P}e^{\mat{\Lambda}(t-s)}\mat{P}^{-1}\mat{D}\mat{Q}e^{\mat{\Phi}(t-s)}\mat{Q}^{-1} \mathrm ds\\
    \intertext{Let $\widetilde{\mat{D}}=\mat{P}^{-1}\mat{D}\mat{Q}$,}
    &=\mat{P}\left(\int_0^t e^{\mat{\Lambda}(t-s)}\widetilde{\mat{D}}e^{\mat{\Phi}(t-s)} \mathrm ds\right)\mat{Q}^{-1}\\
    \intertext{Considering the integral element-wise, we get}
   \int_0^t\left( e^{\mat{\Lambda}(t-s)}\widetilde{\mat{D}}e^{\mat{\Phi}(t-s)}\right)_{ij} \mathrm ds &=\left[-\frac{1}{{\Lambda}_{ii}+\Phi_{jj}} e^{\Lambda_{ii}(t-s)}\widetilde{D}_{ij}e^{\Phi_{jj}(t-s)} \right]_0^t\\
     &=\frac{\widetilde{D}_{ij}}{\Lambda_{ii}+\Phi_{jj}}e^{t(\Lambda_{ii}+\Phi_{jj})}-\frac{\widetilde{D}_{ij}}{\Lambda_{ii}+\Phi_{jj}}
\end{align}
Hence, we get the required result
\begin{equation}
    \label{eq:weight soln final 1}
    \mat{H}(t) = e^{\mat{B}t}\mat{H}(0)e^{\mat{C}t}+ \mat{P}\mat{F}(t)\mat{Q}^{-1},
\end{equation}
where 
\begin{equation}
    F_{ij}(t) = \frac{\widetilde{D}_{ij}}{\Lambda_{ii}+\Phi_{jj}}e^{t({\Lambda}_{ii}+\Phi_{jj})}-\frac{\widetilde{{D}}_{ij}}{{\Lambda}_{ii}+{\Phi}_{jj}}.
\end{equation}
\phantom{QED}\hfill$\Box$
\begin{lemma}
    \label{lemma:2}
    Assuming that $\mat{A}$ and $\mat{W}$ have eigenvalue decompositions $\mat{P}\mat{\Lambda}\mat{P}^{-1}$ and $\mat{Q}\mat{\Phi}\mat{Q}^{-1}$ respectively,
    \begin{equation}
        \int_0^1 \mat{A}^s\mat{E}\mat{W}^s\mathrm ds = \mat{P}\mat{F}\mat{Q}^{-1},
    \end{equation}
    where 
    \begin{equation}
        F_{ij} =\frac{\Lambda_{ii}\widetilde{E}_{ij}\Phi_{jj}-\widetilde{E}_{ij}}{\ln\Lambda_{ii} + \ln \Phi_{jj}},
    \end{equation}
    with $\widetilde{\mat{E}}=\mat{P}^{-1}\mat{E}\mat{Q}$.
\end{lemma}
\textit{Proof}: We start by using the eigenvalue decompositions of $\mat{A}$ and $\mat{W}$
\begin{equation}
    \int_0^1 \mat{A}^s\mat{E}\mat{W}^s\mathrm ds = \mat{P}\int_0^1 \mat{\Lambda}^s\widetilde{\mat{E}}\mat{\Phi}^s\mathrm ds\mat{Q}^{-1},
\end{equation}
where $\widetilde{\mat{E}} = \mat{P}^{-1}\mat{E}\mat{Q}$. Now we can consider the integral element-wise to get the required result:
\begin{align}
    \int_0^1 \left(\mat{\Lambda}^s\widetilde{\mat{E}}\mat{\Phi}^s\right)_{ij}\mathrm ds &= \int_0^1\Lambda_{ii}^s\widetilde{E}_{ij}\Phi_{jj}^s \mathrm ds \\
        &= \left[\frac{\Lambda_{ii}^s\widetilde{E}_{ij}\Phi_{jj}^s}{\ln\Lambda_{ii} + \ln \Phi_{jj}}\right]_0^1\\
        &= \frac{\Lambda_{ii}\widetilde{E}_{ij}\Phi_{jj}-\widetilde{E}_{ij}}{\ln\Lambda_{ii} + \ln \Phi_{jj}}
\end{align}

\phantom{QED}\hfill$\Box$

\subsection{Derivation of the ODE}
\textit{Proof of Proposition 3:}
For the discrete dynamic defined 
\begin{equation}
    \mat{H}_{n + 1} = \mat{A}\mat{H}_n\mat{W} + \mat{H}_0,
\end{equation}
$\mat{H}_n$ can be rewritten as follows:
\begin{equation}
    \mat{H}_n = \sum_{k=0}^n \mat{A}^k \mat{E} \mat{W}^k
\end{equation}
Recall that as in the case where $\mat{W}=\mat{I}$ we move to a continuous setting by interpreting the equation as a Riemann integral:
\begin{equation}
    \mat{H}(t)=\int_{0}^{t+1}\mat{A}^{s}\mat{E}\mat{W}^{s}\mathrm ds.
\end{equation}

To derive a corresponding ODE, we consider the derivative of $\mat{H}(t)$ with respect to $t$, yielding the ODE below:
\begin{equation}
    \frac{\der \mat{H}(t)}{\der t} =  \mat{A}^{t+1}\mat{E}\mat{W}^{t+1}.
\end{equation}
To get the ODE in a nicer form, we consider the second-derivative of $\mat{H}$:
\begin{equation}
   \frac{\der^2 \mat{H}(t)}{\der t^2}(t) =  \ln \mat{A} \mat{A}^{t+1}\mat{E}\mat{W}^{t+1} + \mat{A}^{t+1}\mat{E}\mat{W}^{t+1} \ln \mat{W} = \ln \mat{A} \frac{\der \mat{H}(t)}{\der t} + \frac{\der \mat{H}(t)}{\der t} \ln \mat{W}.
\end{equation}
By integrating over $t$ in both sides of the above equation, we can obtain:
\begin{equation}
    \frac{\der \mat{H}(t)}{\der t}(t) = \ln \mat{A} \mat{H}(t) + \mat{H}(t) \ln \mat{W} + c.
\end{equation}
Note that the initial value of the ODE is $\mat{H}(0)$ by Lemma~2, we know that 
\begin{equation}
    \left(\mat{P}^{-1}\mat{H}(0)\mat{Q}\right)_{ij}= \frac{\Lambda_{ii}\widetilde{E}_{ij}\Phi_{jj}-\widetilde{E}_{ij}}{\ln\Lambda_{ii} + \ln \Phi_{jj}},
\end{equation}
where $\widetilde{\mat{E}}=\mat{P}^{-1}\mat{E}\mat{Q}$.

By setting $t$ to 0, we can obtain:
\begin{equation}
    \left.\frac{\der \mat{H}(t)}{\der t}\right|_{t=0} = \mat{A}\mat{E}\mat{W} \implies \mat{A}\mat{E}\mat{W}-\ln\mat{A}\mat{H}(0)-\mat{H}(0)\ln\mat{W}= c.
\end{equation}
We can simplify $c$ as follows:
\begin{align}
     c &= \mat{A}\mat{E}\mat{W}-\ln\mat{A}\mat{H}(0)-\mat{H}(0)\ln\mat{W}\\
  \left(\mat{P}^{-1}c\mat{Q}\right)_{ij}  &= \Lambda_{ii}\widetilde{E_{ij}}\Phi_{jj} - \ln\Lambda_{ii}\frac{\Lambda_{ii}\widetilde{E}_{ij}\Phi_{jj}-\widetilde{E}_{ij}}{\ln\Lambda_{ii}+\ln\Phi_{jj}}-
    \frac{\Lambda_{ii}\widetilde{E}_{ij}\Phi_{jj}-\widetilde{E}_{ij}}{\ln\mat{\Lambda}_{ii}+\ln\Phi_{jj}}\ln\Phi_{jj}\\
    c &= \mat{P}\widetilde{\mat{E}}\mat{Q}^{-1}\\
    &=\mat{E}
\end{align}
Thus the ODE can be reformulated as:
\begin{equation}
    \frac{\der \mat{H}(t)}{\der t} = \ln \mat{A} \mat{H}(t) + \mat{H}(t) \ln \mat{W} + \mat{E}.
\end{equation}
\phantom{QED}\hfill$\Box$

\textit{Proof of Proposition 4:} Proposition 4 follows trivially from Lemma~\ref{lemma:sylvester}.\\
\phantom{QED}\hfill$\Box$

\section{Hyperparameters \& training details}
Across Cora, Citeseer, and Pubmed we use a hidden dimension of 16, input dropout of 0.5 in the encoder, and weight decay of $5\times10^{-4}$. For NELL we use a hidden dimension of 64, dropout of 0.1 (in both encoder and decoder), and weight decay of $1\times10^{-5}$.$\gamma$ determines the weight of self-loops in the graph.

\FloatBarrier
\begin{table}[h]
    \centering
    \caption{Cora hyperparameters, using the \texttt{rmsprop} optimiser.}
    \label{tab:hyper cora}
    \begin{tabular}{c ccccc ccccc}
    &\multicolumn{5}{c}{fixed split}&\multicolumn{5}{c}{random split}\\
        \cmidrule(lr){2-6} \cmidrule(lr){7-11}
        Algorithm & lr & $t_1$ & $\alpha$ & $\gamma$ & $\beta$ & lr & $t_1$ & $\alpha$ & $\gamma$ &$\beta$\\\hline
         CGNN discrete & $3.45\times10^{-3}$ & $12.0$& $0.950$&$0.309$ &--- & $2.08\times10^{-3}$ & $20.0$ & $0.950$ & $0.517$& ---\\
         CGNN & $4.70\times10^{-3}$ & $12.1$& $0.918$&$0.555$ & --- & $1.47\times10^{-3}$ & $23.9$& $0.885$&$0.595$& --- \\
         CGNN with weights & $2.11\times10^{-3}$ & $14.3$& $0.950$ & $0.947$ & $0.5$&---&---&---&---&$0.5$\\
    \end{tabular}
\end{table}

\begin{table}[h]
    \centering
    \caption{Citeseer hyperparameters, using the \texttt{rmsprop} optimiser.}
    \begin{tabular}{c ccccc ccccc}
        &\multicolumn{5}{c}{fixed split}&\multicolumn{5}{c}{random split}\\
        \cmidrule(lr){2-6} \cmidrule(lr){7-11} Algorithm & lr & $t_1$ & $\alpha$ & $\gamma$ & $\beta$ &lr & $t_1$ & $\alpha$ & $\gamma$& $\beta$ \\\hline
         CGNN discrete & $3.28\times10^{-3}$ & $20.0$& $0.950$&$0.693$ &--- & $2.59\times10^{-3}$ & $20.0$ & $0.950$ & $0.548$& --- \\
         CGNN & $5.48\times10^{-3}$ & $19.1$& $0.869$&$0.758$& --- & $2.98\times10^{-3}$ & $17.1$& $0.936$&$0.459$& ---\\
         CGNN with weights & $5.70\times 10^{-3}$ & $23.4$& $0.950$&$0.775$ &$0.5$&---&---&---&---&$0.5$\\
    \end{tabular}
\end{table}

\begin{table}[h]
    \centering
    \caption{Pubmed hyperparameters, using the \texttt{adam} optimiser.}
    \begin{tabular}{c ccccc ccccc}
    &\multicolumn{5}{c}{fixed split}&\multicolumn{5}{c}{random split}\\
        \cmidrule(lr){2-6} \cmidrule(lr){7-11}
        Algorithm & lr & $t_1$ & $\alpha$ & $\gamma$ & $\beta$ & lr & $t_1$ & $\alpha$ & $\gamma$ & $\beta$\\\hline
         CGNN discrete & $5.57\times10^{-3}$ & $20.0$& $0.950$&$0.626$ &--- & $4.08\times10^{-3}$ & $20.0$ & $0.950$ & $0.933$ & --- \\
         CGNN & $5.40\times10^{-3}$ & $16.2$& $0.960$&$0.644$& ---& $5.51\times10^{-3}$ & $22.0$& $0.947$&$0.752$& ---\\
         CGNN with weights & $5.14\times10^{-3}$ & $20.3$& $0.950$&$0.668$ &$0.5$&---&---&---&---&$0.5$\\
    \end{tabular}
\end{table}

\begin{table}[ht]
    \centering
    \caption{NELL hyperparameters, using the \texttt{adam} optimiser.}
    \begin{tabular}{c ccccc ccccc}
    &\multicolumn{5}{c}{fixed split}&\multicolumn{5}{c}{random split}\\
        \cmidrule(lr){2-6} \cmidrule(lr){7-11}
        Algorithm & lr & $t_1$ & $\alpha$ & $\gamma$ & $\beta$ & lr & $t_1$ & $\alpha$ & $\gamma$&$\beta$\\\hline
         CGNN discrete & $0.01$ & $20.0$ & $0.950$ & $0.941$ &--- & $0.01$ & $20.0$ & $0.950$ & $0.941$& --- \\
         CGNN & $0.01$ &$20.0$ &$0.950$& $0.941$ & --- & $0.01$ &$20.0$ &$0.950$& $0.941$& --- \\
         CGNN with weights & $0.01$ & $20.0$ & $0.950$ & $0.956$ &$0.5$& ---&---&---&---&$0.5$\\
    \end{tabular}
\end{table}

In practice, we used the augmentation proposed in \citet{dupo2019} to stabilise training, however, it had little no effect on the final performance. Writing down the augmentation mathematically yields the following matrix differential equation for the ODE
\begin{equation}
    \frac{\der }{\der t} \begin{bmatrix}
    \mat{H}(t) &\mat{O}(t)
    \end{bmatrix}= (\mat{A}-\mat{I}) \begin{bmatrix}
    \mat{H}(t) &\mat{O}(t)
    \end{bmatrix} + \begin{bmatrix}
    \mat{H}(t) &\mat{O}(t)
    \end{bmatrix} (\mat{W}-\mat{I}) + \begin{bmatrix}
    \mat{E} &\mat{0}
    \end{bmatrix},
\end{equation}
where $\mat{O}(0)=0$.
All the augmentation does is add some latent dimensions to allow the trajectory of the ODE (which cannot cross itself) a potentially simpler path. 

\section{Running time}
In Table~\ref{tab:run time} we compare the running times between our algorithms and GCN as measured on the Cora dataset using the hyperparameters in Table~\ref{tab:hyper cora}. For GCN we used the hyperparameters quoted in \cite{kipf16}. For all models we used 400 epochs and measured the total running time. All results were collected on a single machine with the following specs: Intel Core i7-6700HQ CPU at 2.60GHz with 16GB of RAM with NVIDIA GeForce GTX 960M with 2GB of dedicated GPU memory.
\begin{table}[ht]
    \centering
    \caption{Running time per epoch on the Cora dataset.}
    \label{tab:run time}
    \begin{tabular}{c|ccc}
         Architecture & CGNN & CGNN with weights & GCN  \\ \hline
         Time per epoch (seconds) & 2.18 &2.65& 0.039
 
    \end{tabular}
\end{table}

\section{Comparison to GCN with skip connections}

We compare with GCN with residual links and with 2, 4, 8, and 16 layers respectively. The results on Cora, Citeseer, and PubMed in both the fixed data split setting and random split setting are given in Table~\ref{tab:gcnres}.
\begin{table}[ht]
\centering
    \begin{tabular}{c|cc|cc|cc}
        \multicolumn{1}{c}{}&\multicolumn{2}{c}{Cora}&\multicolumn{2}{c}{Citeseer}&\multicolumn{2}{c}{PubMed}\\
         Number of Layers & Random splits & Fixed splits &Random splits & Fixed splits&Random splits & Fixed splits\\\hline
         2&$77.2\pm1.7$&$77.4\pm1.9$ &$68.0\pm2.5$&$64.7\pm1.5$&   $80.7\pm1.9$&$79.6\pm0.6$\\
         4&$76.5\pm1.6$&$77.3\pm2.0$& $65.6\pm2.3$&$62.6\pm2.0$&  $80.5\pm1.4$&$78.6\pm0.6$\\
         8&$76.8\pm1.9$&$75.7\pm1.3$& $64.2\pm2.1$&$62.3\pm1.3$& $80.0\pm1.4$&$78.5\pm0.9$\\
         16&$77.6\pm2.1$&$76.8\pm1.9$&  $64.4\pm2.0$&$62.2\pm1.9$& $80.4\pm0.3$&$79.9\pm1.3$
    \end{tabular}
    \caption{Performance of GCN with residual links and increasing depth on Cora, Citeseer, and PubMed.}
    \label{tab:gcnres}
\end{table}

\end{document}